\documentclass{article}
\usepackage{fix-cm}

\usepackage{PRIMEarxiv}

\usepackage[utf8]{inputenc} % allow utf-8 input
\usepackage[T1]{fontenc}    % use 8-bit T1 fonts
\usepackage{hyperref}       % hyperlinks
\usepackage{url}            % simple URL typesetting
\usepackage{booktabs}       % professional-quality tables
\usepackage{nicefrac}       % compact symbols for 1/2, etc.
\usepackage{microtype}      % microtypography
\usepackage{dsfont}         % for \mathds
\usepackage{lipsum}
\usepackage{fancyhdr}       % header
\usepackage{graphicx}       % graphics
\graphicspath{{media/}}     % organize your images and other figures under media/ folder
\usepackage{amsmath,amsfonts, amssymb}
\usepackage{dsfont}
\usepackage{booktabs}
\usepackage{algpseudocode, algorithm}
\usepackage{multirow}
\usepackage{array}
\usepackage[caption=false,font=normalsize,labelfont=sf,textfont=sf]{subfig}
\usepackage{textcomp}
\usepackage{stfloats}
\usepackage{url}
\usepackage{verbatim}
\usepackage{graphicx}
\title{FlashReg: GPU-Accelerated 3-Clique Point Cloud Registration for Real-Time Correspondence-to-Pose Estimation
}

\author{
    Ziyang Yu\\
    School of Computing \\
    Institute of Science Tokyo\\
    Tokyo \\
    \texttt{yu.ziyang.2001@gmail.com}
    \And
    Xiang Li \\
    School of Computing \\
    Institute of Science Tokyo \\
    Tokyo \\
    \texttt{652022230010@smail.nju.edu.cn}
    \And
    Qiong Chang \\
    School of Computing \\
    Institute of Science Tokyo \\
    Tokyo \\
    \texttt{q.chang@c.titech.ac.jp}
    \And
    Jun Miyazaki \\
    School of Computing \\
    Institute of Science Tokyo \\
    Tokyo\\
    \texttt{miyazaki@comp.isct.ac.jp}
}

\begin{document}
\maketitle

\begin{abstract}
Graph-based point cloud registration achieves high robustness by identifying geometrically consistent correspondence sets, but constructing second-order compatibility graphs and enumerating candidate cliques remain compute- and memory-intensive. This work presents FlashReg, a GPU-oriented correspondence-to-pose estimator that avoids materializing the dense scored second-order graph. Its Fast First- and Second-Order Graph (FFSOG) construction builds a capacity-bounded sparse second-order graph directly from the binary first-order graph. A dataflow-optimized three-node clique (3-clique) search then selects pivots from compact per-row candidate pools and enumerates triples through sorted sparse-neighborhood intersections. Across indoor and outdoor benchmarks, FlashReg reduces correspondence-to-pose latency by 2--3$\times$ relative to TurboReg at comparable registration recall, while using about 50\% of its peak allocated tensor memory on an embedded GPU. These results make FlashReg suitable as a high-throughput registration backend within onboard perception pipelines.
\end{abstract}

% keywords can be removed

\section{Introduction}
% 第1段（背景与问题）：点云配准的目标是将两个独立点云在同一坐标系下对齐的过程，通常用于机器人，SLAM，自动驾驶，虚拟现实等多个场景。由于点云信息存在稀疏和旋转不变的特点，相比于二维图像，点云配准一直存在可用特征少，数据相似度低，配准速度慢的问题。然而，在绝大多数的点云配准应用中，通常对于实时性和配准质量有着同样高的要求，因此设计高精度高效率的点云配准算法具有重要的现实意义。
% \IEEEPARstart{P}{OINT} 
Point cloud registration recovers the rigid transformation that aligns two independently captured scans into a common coordinate frame, which underpins robot perception, SLAM, autonomous driving, and virtual reality~\cite{pcregreview,loam,kissicp}. Compared with 2D images, point clouds are sparse and carry no canonical orientation, so they expose far fewer distinctive features and far lower cross-scan similarity; registration has therefore long been burdened by high computational cost and memory requirements. Most real deployments, however, demand real-time response and high registration quality at the same time, especially on the resource-limited platforms where these algorithms actually deploy.

% 第2段：直接从 graph-based registration 切入，概括其鲁棒性、计算瓶颈及 TurboReg 尚未解决的稠密数据流问题。
Graph-based registration represents each putative correspondence as a node and connects pairs that satisfy pose-invariant geometric consistency~\cite{spectralmatching,teaser++,sc2pcr}. Cliques in this graph isolate self-consistent correspondence sets, while second-order compatibility further suppresses spurious edges, enabling strong robustness under high outlier rates~\cite{sc2pcr,mac,mac++}. This accuracy is expensive: graph construction grows quadratically with the correspondence count, and maximal-clique search may enumerate millions of overlapping candidates. TurboReg~\cite{turboreg} replaces exhaustive enumeration with GPU-based pivot search, extending each high-scoring second-order edge into a bounded set of 3-cliques and achieving more than a 100$\times$ speedup over CPU-based MAC. However, it still materializes dense compatibility matrices and rebuilds intermediate results between stages. The resulting computation and memory traffic are particularly costly on unified-memory embedded GPUs.

\begin{figure}[t]
    \centering
    \includegraphics[width=0.55\linewidth]{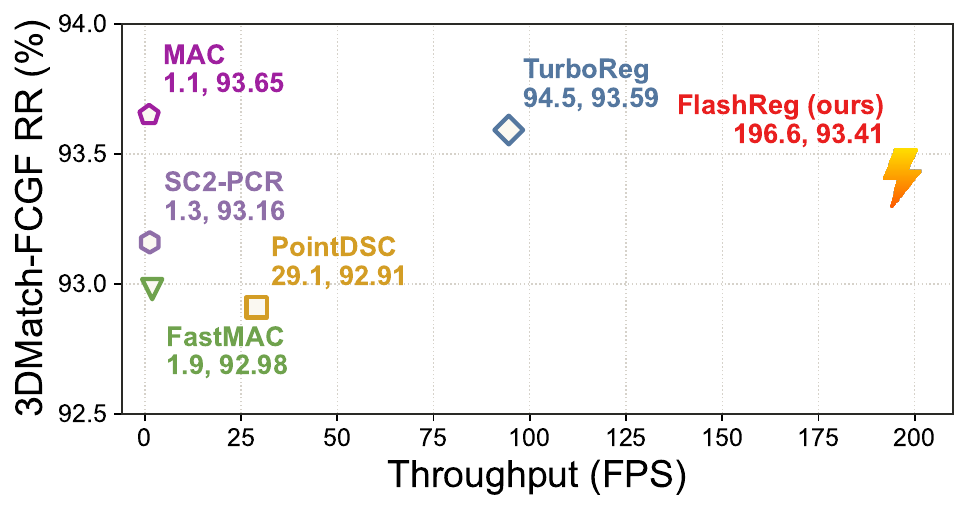}
    \caption{Speed--accuracy trade-off on 3DMatch with FCGF descriptors; each label gives estimator throughput (FPS) and recall (\%), and the truncated recall axis magnifies sub-point differences. FlashReg stays within half a point of the strongest evaluated solvers at nearly 200\,FPS.}
    \label{fig:speed_recall}
\end{figure}

% 第3段：在明确继承 TurboReg 3-clique 目标的基础上，聚焦 FlashReg 如何消除稠密存储与重复计算。
Building on the 3-clique formulation, we present \textbf{FlashReg}, a clique-based registration framework that retains TurboReg's pivot-based 3-clique scoring objective while using a capacity-bounded candidate set. Our design is guided by two structural observations. First, the graph is sparse: each correspondence is compatible with only a small fraction of the others, making a dense representation unnecessary. Second, graph construction already produces much of the information subsequently needed for pivot selection and clique enumeration. By co-designing graph construction and inter-stage data movement at the kernel level, these intermediate results can be passed directly to the following processes rather than recomputed, substantially reducing the computational and memory overhead of clique enumeration.

% 第4段：两个设计（机制细节留给 Sec. III 与贡献列表）。
These observations motivate two designs in FlashReg: (1) Fast First- and Second-Order Graph (FFSOG) construction, which computes and compacts SOG rows using on-chip buffers; and (2) a dataflow-optimized 3-clique search, which fuses pivot selection with SOG construction and enumerates cliques directly from the compact sparse representation, avoiding the write-back of a dense SOG score matrix. At the cost of a slightly more elaborate construction kernel, FlashReg removes most of the memory traffic and computation of the two most expensive stages of the pipeline.
As illustrated in Fig.~\ref{fig:speed_recall}, FlashReg stays within half a recall point of the strongest evaluated solvers at an estimator throughput of nearly \textbf{200\,FPS}. The savings are larger still on embedded GPUs: FlashReg uses about 50\% of TurboReg's peak allocated tensor memory and consumes less measured GPU-rail energy at comparable registration recall. FlashReg thereby turns accurate 3-clique registration from a dense, stage-separated pipeline into a compact, memory-efficient backend suitable for resource-constrained robot platforms. The main contributions are summarized as follows:
% 贡献列表：1）GPU 原生稀疏图构建；2）基于数据流向优化的最大团搜索算法；3）实验。
\begin{itemize}
\item A compact, GPU-native graph construction method that removes the dense SOG score-matrix bottleneck from 3-clique registration. FFSOG fuses first-order compatibility evaluation and constructs the scored SOG directly in a capacity-bounded sparse layout, avoiding materialization of the dense score matrix while retaining comparable registration recall.

\item A dataflow-optimized 3-clique search algorithm that co-designs pivot selection and clique enumeration with graph construction. It selects pivots directly from row-local candidates produced during SOG construction and enumerates cliques through sorted sparse-neighborhood intersections, avoiding dense intermediate storage, redundant computation, and host synchronization.

\item Experiments on indoor, outdoor, and low-overlap benchmarks show that FlashReg preserves competitive recall while reducing estimator latency by more than $2\times$; on the Jetson AGX Xavier, it uses about 50\% of TurboReg's peak allocated tensor memory with lower measured GPU-rail energy.
\end{itemize}

\section{Related Work}
\subsection{Descriptor-based Point Cloud Registration}
% 经典的点云配准以ICP~\cite{icp}为代表，通过迭代最近点不断优化刚性变换，但ICP强烈依赖良好的初始位姿，容易陷入局部最优，难以处理无初值的全局配准。为摆脱对初值的依赖，研究者转向基于描述子的配准：先在两个点云上分别计算局部几何特征，再在特征空间中匹配，从而无需初始位姿即可建立点对点的对应关系。早期为手工描述子，如Spin Image~\cite{spinimage}、PFH~\cite{pfh}、FPFH~\cite{fpfh}、SHOT~\cite{shot}，依据局部法向与几何统计刻画邻域；近年来随着深度学习发展，出现了3DMatch~\cite{3dmatch}、FCGF~\cite{fcgf}等学习式描述子，在鲁棒性和区分度上明显更强；进一步地，Predator~\cite{predator}、GeoTransformer~\cite{geotransformer}等方法引入重叠感知与注意力匹配，直接学习跨扫描的对应关系。然而，受点云稀疏性与旋转不变性的限制，即便是最好的描述子，在低重叠、高噪声场景下匹配得到的对应关系中仍混有大量外点，因此对于外点的筛选过滤能力是点云配准算法的重要课题。
ICP~\cite{icp} requires a good initial pose, while globally optimal variants such as Go-ICP~\cite{goicp} remain far from real time. Descriptor-based methods avoid initialization by matching handcrafted~\cite{spinimage,pfh,fpfh,shot} or learned~\cite{3dmatch,3dsmoothnet,fcgf} local features; Predator~\cite{predator} and GeoTransformer~\cite{geotransformer} further add overlap-aware matching. Even so, low-overlap and noisy scans leave many outliers, making robust filtering essential.

\subsection{Graph-based Point Cloud Registration}
% 为在含大量外点的对应集合中稳健地筛出内点，研究者提出了基于图（兼容性）的配准：把每个对应作为节点，若两个对应在距离/角度等度量上相互兼容则连边；由于成对兼容性对未知位姿保持不变，因此一组两两兼容的对应（团）天然构成一个自洽的内点集合。外点剔除大体分为两条路线。一类是学习式方法：PointDSC~\cite{pointdsc}用深度网络学习空间一致性来给对应打分，DGR~\cite{dgr}以卷积网络分类内点并加权求解位姿，3DRegNet~\cite{3dregnet}则直接从对应回归位姿；这类方法精度高，但依赖有标注数据训练，跨传感器、跨场景的泛化能力受限。另一类是免训练的几何方法，直接利用成对兼容性构图并做团搜索，无需训练即可跨域使用，本文即沿这一路线。TEASER++~\cite{teaser++}沿这一思路，用平移不变量(TIM)/旋转不变量(TRIM)构建兼容性图，先用最大团做内点剪枝，再以图优化(GNC)稳健求解位姿，对高外点率具有很强的鲁棒性。SC2-PCR~\cite{sc2pcr}进一步引入二阶空间兼容性：在一阶兼容的基础上，要求两个对应共同兼容的邻居也相互兼容，以此更严格地收紧candidate范围，显著抑制了外点带来的误连边。MAC~\cite{mac}指出以往方法追求单个最大团(maximum clique)会丢弃大量有效内点，转而枚举所有极大团(maximal clique)并放宽约束，从每个团分别估计位姿再择优，从而在低重叠场景下取得了当时最好的精度。但精度的代价是效率：兼容性图的规模随对应数二次增长，构图时逐对过滤外点开销巨大，且有限的数据往往生成千万量级、结构高度相似的极大团，团搜索需在其中反复筛选——这使得这类方法虽准却慢。
Graph-based registration represents correspondences as nodes and connects pose-invariant compatible pairs~\cite{spectralmatching}; a clique then forms a self-consistent inlier set. Learning-based filters such as PointDSC~\cite{pointdsc}, DGR~\cite{dgr}, and 3DRegNet~\cite{3dregnet} require labeled training, whereas geometric methods search compatibility graphs directly. TEASER++~\cite{teaser++} combines clique pruning with GNC, and SC2-PCR~\cite{sc2pcr} uses second-order compatibility to suppress spurious edges. MAC~\cite{mac} loosens clique-enumeration constraints and estimates a pose per clique, improving low-overlap accuracy. This robustness introduces two bottlenecks: graph construction evaluates and stores $O(N^2)$ pairwise relations, while maximal-clique enumeration can generate millions of overlapping candidates that must be scored and converted into pose hypotheses.

\subsection{Registration Algorithm Acceleration}
% 早期的配准加速集中在CPU端的算法优化，利用点云自身的结构信息削减重复计算，如FGR~\cite{fgr}以一次性优化的鲁棒目标替代反复的对应搜索与迭代，避开了ICP式的迭代开销。随着GPU算力提升，研究者开始把配准迁移到GPU；但早期GPU加速多针对结构简单、天然并行的几何/描述子类算法（如特征提取、最近邻匹配），而基于图/团的方法因团枚举带有递归与强数据依赖，与GPU的大规模并行模型天然冲突，长期难以有效并行。针对团配准，出现了两条加速路线。其一是降规模：FastMAC~\cite{fastmac}用图信号采样在搜索前稀释对应集合，降低了绝对开销，但并未触及组合爆炸的本质，采样率一高精度就明显下滑。其二是换硬件：TurboReg~\cite{turboreg}首次将团配准的完整流程搬上GPU，通过收紧团的搜索范围提出基于顶点(pivot)的高效搜索，相较CPU端的MAC取得百倍以上加速，证明了GPU团配准的可行性。然而TurboReg沿用面向服务器GPU的稠密算子，既未充分利用GPU的细粒度并行，又在计算与显存上存在明显冗余；在显存/供电受限的嵌入式统一内存平台上，这些冗余会直接转化为不必要的内存与能耗开销。因此，如何在GPU上进一步压榨团配准的效率、做到更快更省，仍是一个亟待解决的问题——这正是FlashReg的出发点。
General acceleration methods optimize CPU solvers~\cite{gicp,vanicp,vanicpgpu,fgr} or parallel descriptor workloads on GPUs, but recursive clique search and its data dependencies remain difficult to parallelize. FastMAC~\cite{fastmac} samples the graph before enumeration to reduce candidates, but aggressive sampling degrades accuracy. TurboReg~\cite{turboreg} instead ports the pipeline to the GPU and replaces maximal-clique enumeration with pivot-based 3-clique search. Since three non-collinear correspondences determine a rigid transformation, it achieves over 100$\times$ speedup versus CPU-based MAC. However, TurboReg still materializes the dense SOG score matrix, applies dense row operations, and rebuilds intermediate results between stages. FlashReg retains TurboReg's search objective while addressing these remaining graph-construction and data-movement costs through sparse SOG construction and a coordinated GPU dataflow.

\begin{figure}
    \includegraphics[width=\columnwidth]{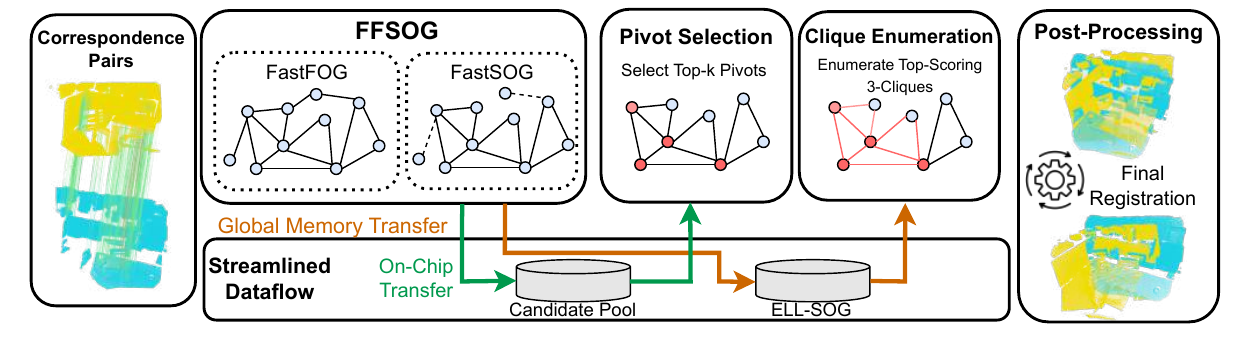}
    \caption{FlashReg pipeline. FastFOG and FastSOG form FFSOG, which emits column-sorted ELL-SOG rows and per-row top-$m$ candidate pools without materializing the dense SOG score matrix. The dataflow-optimized search consumes these outputs in place. Green and orange arrows denote on-chip and global-memory traffic, respectively.}
    \label{fig:flashreg_pipeline}
\end{figure}

\section{Method}
% ── Method 开头（问题设定 + 流水线 + 记号，原 Overview and Notation 小节已并入）──

Given a source and a target point cloud, descriptor matching yields a set of $N$ putative correspondences $\mathcal{C}=\{c_i\}_{i=1}^{N}$, where each $c_i=(p_i^s,p_i^t)$ pairs a source point $p_i^s$ with a target point $p_i^t$. Registration seeks the rigid transformation that aligns the inlier correspondences,
\begin{equation}
(\mathbf{R}^\star,\mathbf{t}^\star)=\arg\min_{\mathbf{R}\in SO(3),\,\mathbf{t}\in\mathbb{R}^3}\ \sum_{i\in\mathcal{I}}\big\|\mathbf{R}\,p_i^s+\mathbf{t}-p_i^t\big\|_2^2,
\label{eq:objective}
\end{equation}
where $SO(3)$ denotes the group of 3D rotations ($\mathbf{R}^\top\mathbf{R}=\mathbf{I}$, $\det\mathbf{R}=1$), and $\mathcal{I}$ denotes the unknown subset of inlier correspondences. FlashReg searches for geometrically consistent 3-cliques to generate pose hypotheses, which are subsequently verified against all correspondences.

As illustrated in Fig.~\ref{fig:flashreg_pipeline}, FlashReg organizes second-order-graph clique registration into four stages: FastFOG, FastSOG, pivot selection, and clique enumeration. FastFOG and FastSOG together form FFSOG (Sec.~\ref{sec:ffsog}), while the dataflow-optimized search (Sec.~\ref{sec:dataflow}) covers the latter two stages. The retained 3-cliques become pose hypotheses, which are verified and refined as described in Sec.~\ref{sec:pose}.

\subsection{FFSOG: Fast First- and Second-Order Graph Construction}
\label{sec:ffsog}
\begin{figure*}[t]
    \centering
    \includegraphics[width=0.84\textwidth]{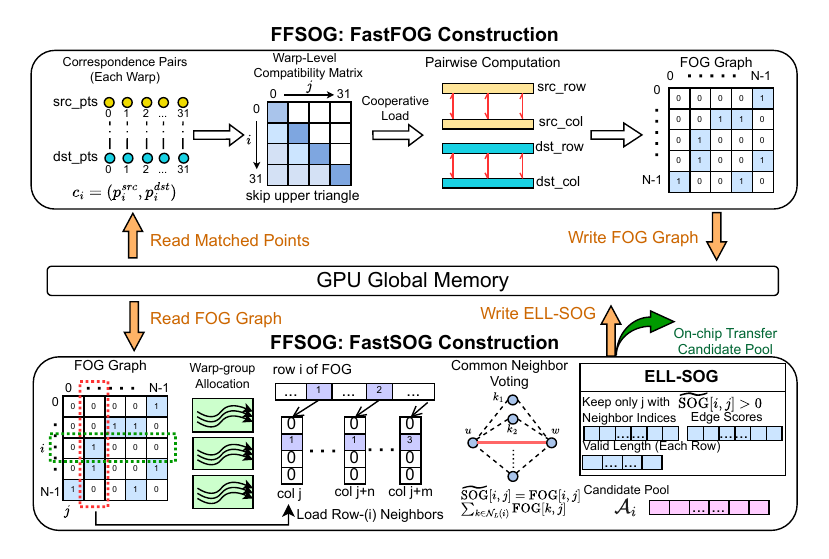}
    \caption{FFSOG. Top, FastFOG: a fused, tiled kernel builds the FOG adjacency matrix and writes it to global memory once. Bottom, FastSOG: one block per node computes a row of the SOG score matrix in shared memory by common-neighbor voting and compacts the nonzeros via warp ballot and prefix sum; before the row leaves shared memory, its $m$ highest-scoring retained edges are extracted into the compact pivot-candidate pool $\mathcal{A}_i$. }
    \label{fig:ffsog}
\end{figure*}

Fig.~\ref{fig:ffsog} illustrates the two modules of FFSOG. The upper FastFOG path builds the first-order graph (FOG) with a single fused, tiled kernel and writes its adjacency matrix to global memory once. The lower FastSOG path converts the FOG into the weighted second-order graph (SOG) row by row: one thread block gathers a node's FOG neighbors into shared memory, computes its SOG edge scores by common-neighbor voting, and emits the positive entries directly as a column-sorted sparse adjacency, the ELL-SOG, alongside a per-row pool of the top-scoring edges for the subsequent pivot selection. FlashReg retains the dense binary FOG adjacency matrix for first-order lookups. The larger dense SOG score matrix, however, is never materialized. The following sections describe each module; rows that exceed the configured capacity fall back to a capped approximation, quantified in Secs.~\ref{sec:dataflow} and~\ref{sec:ablation}.

\subsubsection{FastFOG Construction}
First-order compatibility tests whether a pair of correspondences preserves its intra-cloud distance. To distinguish a graph from its representation, we use FOG and SOG for the graphs and $\mathbf{M}_{\mathrm{FOG}}$ and $\mathbf{M}_{\mathrm{SOG}}$ for their matrices. In particular, let $\mathcal{G}_{\mathrm{FOG}}=(\mathcal{C},\mathcal{E}_{\mathrm{FOG}})$ denote the undirected FOG and let $\mathbf{M}_{\mathrm{FOG}}\in\{0,1\}^{N\times N}$ denote its symmetric adjacency matrix.

As shown in the upper path of Fig.~\ref{fig:ffsog}, FastFOG evaluates the compatibility tests with one fused, tiled kernel. A $32\times32$ block produces one tile of $\mathbf{M}_{\mathrm{FOG}}$, with each warp comparing one row correspondence against 32 column correspondences. FastFOG evaluates only the lower-triangular half of the matrix and writes every off-diagonal result to both $[\mathbf{M}_{\mathrm{FOG}}]_{ij}$ and $[\mathbf{M}_{\mathrm{FOG}}]_{ji}$. It therefore produces the complete dense binary adjacency matrix; FastSOG later stores only the strict upper-triangular positive SOG entries. Let $i_0$ and $j_0$ be the tile origins; for local indices $a,b\in\{0,\ldots,31\}$, the block cooperatively stages the row points $(p^s_{i_0+a},p^t_{i_0+a})$ and column points $(p^s_{j_0+b},p^t_{j_0+b})$ in shared memory (Fig.~\ref{fig:ffsog}, top). After synchronization, thread $(a,b)$ reuses these arrays with $i=i_0+a$ and $j=j_0+b$, and computes
\begin{equation}
\ell^s_{ij}=\big\|p_i^s-p_j^s\big\|_2,\quad
\ell^t_{ij}=\big\|p_i^t-p_j^t\big\|_2,\quad
d_{ij}=\big|\ell^s_{ij}-\ell^t_{ij}\big|,
\label{eq:dis}
\end{equation}
followed by the binary compatibility score
\begin{equation}
[\mathbf{M}_{\mathrm{FOG}}]_{ij}
=\mathds{1}(i\ne j)\,
 \mathds{1}\!\left(d_{ij}<\tau\right)\,
 \mathds{1}\!\left(\ell^s_{ij}+\ell^t_{ij}>r\right),
\label{eq:fog}
\end{equation}
where $\mathds{1}(\cdot)$ is the indicator function, which equals one when its argument is true and zero otherwise. Thus, $(i,j)\in\mathcal{E}_{\mathrm{FOG}}$ exactly when $[\mathbf{M}_{\mathrm{FOG}}]_{ij}=1$. Both $\ell^s_{ij}$ and $\ell^t_{ij}$ are Euclidean distances, so the length-consistency threshold $\tau$ and NMS radius $r$ use the same length unit as the point coordinates.

Compared with stage-by-stage construction, FastFOG fuses all compatibility operations into one kernel, reuses tiled coordinates in shared memory, and computes only one triangle of the symmetric matrix. It produces the same FOG adjacency matrix with $O(N^2)$ computation and storage while reducing constant-factor computation and global-memory traffic, cutting construction time from $1.64$ to $0.19$\,ms (Sec.~\ref{sec:ablation}).

\subsubsection{FastSOG Construction}
FastSOG is designed around a mismatch between SOG computation and its downstream use. The original method materializes an $N\times N$ scored matrix. General-purpose sparse matrix multiplication incurs metadata construction, symbolic discovery, and output allocation that are unnecessary for our fixed consumer-oriented layout. FastSOG instead exploits the binary masked-voting structure to score and compact each row directly into the two representations consumed downstream.

Let $\mathbf{M}_{\mathrm{SOG}}\in\mathbb{R}_{\ge 0}^{N\times N}$
denote the symmetric score matrix of the weighted SOG. It is computed directly from the binary FOG adjacency matrix by a masked matrix product,
\begin{equation}
\begin{split}
\mathbf{M}_{\mathrm{SOG}}
&=\mathbf{M}_{\mathrm{FOG}}\odot
\left(\mathbf{M}_{\mathrm{FOG}}\times\mathbf{M}_{\mathrm{FOG}}\right),\\
[\mathbf{M}_{\mathrm{SOG}}]_{ij}
&=[\mathbf{M}_{\mathrm{FOG}}]_{ij}\sum_{k=1}^{N}
[\mathbf{M}_{\mathrm{FOG}}]_{ik}
[\mathbf{M}_{\mathrm{FOG}}]_{kj},
\end{split}
\label{eq:sog}
\end{equation}
where $\odot$ denotes element-wise product and $\times$ denotes matrix product. The matrix-product term counts the common FOG neighbors shared by $i$ and $j$. To describe this neighborhood structure, let $\mathcal{N}_{\mathrm{FOG}}(i)=\{k:[\mathbf{M}_{\mathrm{FOG}}]_{ik}=1\}$ denote the FOG neighborhood of node $i$, with degree $d_i=|\mathcal{N}_{\mathrm{FOG}}(i)|$. In Eq.~\eqref{eq:sog}, the outer factor $[\mathbf{M}_{\mathrm{FOG}}]_{ij}$ forces the SOG score to zero whenever correspondences $i$ and $j$ are not connected in the FOG. The SOG therefore never introduces a new edge; it only assigns a common-neighbor score to an existing FOG edge. Consequently, row $i$ of $\mathbf{M}_{\mathrm{SOG}}$ can contain at most $d_i$ nonzero entries, and only neighbors $k\in\mathcal{N}_{\mathrm{FOG}}(i)$ can contribute to their scores. A dense matrix multiplication nevertheless evaluates all $N^2$ output cells using length-$N$ inner products. With mean FOG degree $\bar d\ll N$, at most a fraction $\bar d/N$ of these cells can carry useful information.

FastSOG exploits both levels of sparsity in the lower path of Fig.~\ref{fig:ffsog}. One thread block handles each row $i$, scans row $i$ of $\mathbf{M}_{\mathrm{FOG}}$ to obtain its candidate columns $\mathcal{N}_{\mathrm{FOG}}(i)$, and retains up to $L$ of them as a shared-memory voter list $\mathcal{N}_{L}(i)\subseteq\mathcal{N}_{\mathrm{FOG}}(i)$; when $d_i>L$, the list keeps the first $L$ neighbors encountered by the block's strided row scan, an implementation-defined subset. The expensive common-neighbor vote is evaluated only for $j\in\mathcal{N}_{\mathrm{FOG}}(i)$ with $j>i$:
\begin{equation}
[\widetilde{\mathbf{M}}_{\mathrm{SOG}}]_{ij}
=[\mathbf{M}_{\mathrm{FOG}}]_{ij}
\sum_{k\in\mathcal{N}_{L}(i)}
[\mathbf{M}_{\mathrm{FOG}}]_{kj}.
\label{eq:sog_trunc}
\end{equation}
Here, $\widetilde{\mathbf{M}}_{\mathrm{SOG}}$ denotes FlashReg's capacity-bounded approximation of the exact score matrix $\mathbf{M}_{\mathrm{SOG}}$. Each retained neighbor $k$ contributes one vote exactly when it also connects to $j$. The initial row scans cost $O(N^2)$, while voting costs $O(\sum_i d_i\min(d_i,L))$, eliminating the cubic dense matrix-product work; the overall graph pipeline still includes FastFOG's unavoidable quadratic pairwise evaluation. The scores remain in registers or shared memory and are compacted in increasing column order by a warp ballot and prefix sum. To make the output layout explicit, let $j_{i,1}<\cdots<j_{i,z_i}$ be the $z_i$ columns satisfying $j>i$ and $[\widetilde{\mathbf{M}}_{\mathrm{SOG}}]_{ij}>0$, let $\ell_i=\min(z_i,L)$, and write $s_{i,t}=[\widetilde{\mathbf{M}}_{\mathrm{SOG}}]_{i,j_{i,t}}$. The three rows shown in the ELL-SOG box of Fig.~\ref{fig:ffsog} are
\begin{equation}
\begin{aligned}
\text{Neighbor Indices}_i
&=[j_{i,1},\ldots,j_{i,\ell_i},
  \underbrace{\ast,\ldots,\ast}_{L-\ell_i}],\\
\text{Edge Scores}_i
&=[s_{i,1},\ldots,s_{i,\ell_i},
  \underbrace{\ast,\ldots,\ast}_{L-\ell_i}],\\
\text{Valid Length}_i&=\ell_i .
\end{aligned}
\label{eq:ell_sog}
\end{equation}
Here, $\ast$ denotes padding ignored by downstream kernels. The fixed-width padded rows, together with an explicit per-row valid length, form an ELLPACK-R layout~\cite{ellpackr}; we term the resulting structure the \emph{ELL-SOG}. Its fixed row stride keeps every row directly addressable by index on the GPU, while storing valid entries in ascending \emph{column} order enables the binary-search intersection described in Sec.~\ref{sec:dataflow}. Before a compacted row leaves shared memory, the block additionally extracts its $m$ highest-scoring edges ($m{=}32$) into a per-row \emph{candidate pool} $\mathcal{A}_i$, ordered by descending score. Thus, the ELL-SOG supports clique enumeration through column-ordered access, whereas $\mathcal{A}_i$ supports pivot selection through score-ordered access (Fig.~\ref{fig:ffsog}, bottom right).

Compared with the original method, FastSOG avoids both the $O(N^2)$ dense SOG score-matrix storage and the later dense-row scans. Its ELL-SOG and candidate pools are generated during compaction in the exact access orders needed by enumeration and pivot selection, respectively. The scored graph therefore occupies $O(NL)$ space, and downstream stages neither reconstruct an adjacency structure nor re-sort its edges. The dense binary FOG adjacency matrix is still retained for first-order lookups.

% ── 以下为实现备注（开发用，不渲染）──
% FFSOG 的目标：FastFOG 用一个融合 tiled kernel 一次建好稠密 FOG 写回显存（build_c2_hard；唯一消费者是 SOG 构建，读完即释放）；
% FastSOG 从显存按行读 FOG 邻居，SOG 全程在 shared memory 构建并压缩，稠密 SOG 一次都不写回显存。具体加速步骤：
% 1) 一阶图 FOG：FOG[i,j] 只由两条对应的坐标决定
%    (dis_ij = | ||src_i-src_j|| - ||dst_i-dst_j|| |，判 tau 与 NMS)，tile 坐标载入 shared memory 后单 kernel 算完，
%    对称镜像只算下三角。注意：代码里存在 FOG 即用即算不落地的全融合变体 fused_c2_sc2_pipeline（fused_c2_sc2_pivot），
%    但未接入任何入口且实测不如当前路径，论文不描述它。
% 2) 一个 block 负责一个节点 i：把节点 i 的一阶邻居 (s_neigh/s_weight) 载入 shared memory；
%    得分累加与稀疏压缩状态留在片上，但共同邻居投票仍按需读取 FOG[k,j]，稠密 SOG 不写回显存。
% 3) SOG 得分即"共同邻居数"，是 [0, K_MAX] 内的小整数：这既让打分变成一次整数计数，
%    又为后面 pivot 选择的"精确直方图桶选"埋下伏笔（见 Dataflow）。
% 4) 稀疏压缩用 warp 原语，不用原子扫描：对本行候选列做 __ballot_sync 得到 valid 掩码，
%    warp 内 prefix-sum + warp 间基址，做 block-stable 紧凑，把非零元直接写进按列升序的 padded/CSR 邻接。
% 5) 只存严格上三角 (col > u)：3-clique 搜索只需要上三角，每条边只存一次，存储与计算都减半。
% 6) 产物即下一阶段所需布局：col_padded/val_padded/row_nnz（团搜索用）与 per-row top-m 候选池
%    cand_score/cand_col [N, TOPM]（pivot 用）一次性产出，稠密 N×N 的 SOG 从未进入全局显存。

\subsection{Dataflow-Optimized 3-Clique Search}
\label{sec:dataflow}

Building on the two complementary representations produced by FFSOG, the search stage follows the pivot-based 3-clique objective of TurboReg~\cite{turboreg} to generate a bounded set of geometrically consistent correspondence triples. Given a pivot budget $K$ and a per-pivot clique budget $T$, Phase~1 selects the $K$ highest-scoring edges from the candidate union $\widetilde{\mathcal{E}}=\bigcup_i\mathcal{A}_i$ as \emph{pivots}. This candidate-restricted selection approximates the global top-$K$ over all positive SOG edges, as evaluated in Sec.~\ref{sec:ablation}, and records each pivot as $(u,v,s_{uv})$. Phase~2 intersects the column-sorted ELL-SOG neighborhoods of each pivot's endpoints, scores their common neighbors $w$, and retains at most the $T$ highest-scoring 3-cliques $(u,v,w)$ for each pivot. We use $T{=}2$ in all experiments. The search therefore emits at most $KT$ scored triples; because only the strict upper triangle is stored, every emitted triple satisfies $u<v<w$ and cannot be generated twice from different pivots.

\subsubsection{Search Objective and Data-Access Mismatch} Let $\mathbf{M}$ denote the SOG score matrix supplied to the search: $\mathbf{M}=\mathbf{M}_{\mathrm{SOG}}$ for the dense reference, while FlashReg uses the retained entries of $\widetilde{\mathbf{M}}_{\mathrm{SOG}}$ stored in ELL-SOG. Define $\mathcal{N}^{+}(i)=\{j>i:[\mathbf{M}]_{ij}>0\}$ as the upper-triangular SOG neighborhood of node $i$ and $S(i,j)=[\mathbf{M}]_{ij}$ as the weight of edge $(i,j)$. The search first selects the $K$ highest-scoring positive upper-triangular edges as pivots. Formally,
\begin{equation}
\mathcal{P}=
\operatorname*{top\text{-}K}_{u<v,\ S(u,v)>0}
S(u,v).
\label{eq:pivot}
\end{equation}
The dense reference breaks ties at the $K$-th score by index. For each pivot $(u,v)\in\mathcal{P}$, the admissible third vertices are their common upper-triangular neighbors:
\begin{equation}
\mathcal{W}(u,v)=\mathcal{N}^{+}(u)\cap\mathcal{N}^{+}(v).
\label{eq:common}
\end{equation}
Each resulting triple receives the sum of its three edge scores:
\begin{equation}
q(u,v,w)=S(u,v)+S(u,w)+S(v,w).
\label{eq:clique}
\end{equation}
The search retains only the $T$ highest-scoring triples for each pivot.

A conventional implementation materializes the dense $N\times N$ SOG score matrix. Pivot selection then examines $O(N^2)$ cells, while clique enumeration scans two length-$N$ rows for each pivot and touches $O(KN)$ entries. This is wasteful because Eqs.~\eqref{eq:pivot}--\eqref{eq:clique} require only positive edges, their scores, and intersections between sparse neighborhoods.

The ELL-SOG exposes exactly this information. For row $i$, let $\mathbf{J}_i$ and $\mathbf{Q}_i$ denote its valid neighbor indices and corresponding edge scores. Then
\begin{equation}
\begin{aligned}
\mathcal{N}^{+}(i)&=\{\mathbf{J}_i[t]:1\le t\le\ell_i\},\\
S(i,\mathbf{J}_i[t])&=\mathbf{Q}_i[t],
\end{aligned}
\label{eq:ell_lookup}
\end{equation}
for $1\le t\le\ell_i$. Downstream kernels inspect only this valid prefix and never read the padding. Alg.~\ref{alg:clique} can therefore evaluate Eqs.~\eqref{eq:pivot}--\eqref{eq:clique} without scanning a dense row or reconstructing an adjacency representation. The pseudocode uses one-based array indices and zero-based thread identifiers.

\begin{algorithm}[h]
\caption{Dataflow-Optimized 3-Clique Search}
\label{alg:clique}
\footnotesize
\begin{algorithmic}[1]
\Require ELL-SOG rows $(\mathbf{J}_i,\mathbf{Q}_i,\ell_i)$ and lookup $S$ from Eq.~\eqref{eq:ell_lookup}
\Require per-row top-$m$ candidate pools $\{\mathcal{A}_i\}$; pivot budget $K$; per-pivot clique budget $T$; threads per block $B$; per-thread register-list capacity $\kappa{\ge}T$
\Ensure up to $T$ highest-scoring valid 3-cliques for each pivot
\Statex \hspace{-1.2em}\textit{Phase 1: dataflow-integrated pivot selection}
\State build histogram $H$ over the integer scores $S(e)$ of all pool edges $e\in\bigcup_i\mathcal{A}_i$
\State $\theta\gets$ the $K$-th largest score among $e\in\bigcup_i\mathcal{A}_i$ \Comment{scan $H$ top-down}
\State $\mathcal{P}\gets\{e:S(e)>\theta\}$; fill from $\{e:S(e){=}\theta\}$ until $|\mathcal{P}|=K$; record each pivot with its pool score as $(u,v,s_{uv})$
\Statex \hspace{-1.2em}\textit{Phase 2: 3-clique enumeration (one block per pivot)}
\ForAll{pivot $(u,v,s_{uv})\in\mathcal{P}$ \textbf{in parallel}}
    \For{thread $t=0,\dots,B{-}1$ \textbf{in parallel}}
        \State $R_t\gets\{(-\infty,\varnothing)\}^{\kappa}$ \Comment{thread-local top-$\kappa$ in registers}
        \For{$a=t+1$ \textbf{to} $\ell_u$ \textbf{step} $B$}
            \State $w\gets\mathbf{J}_u[a]$
            \State $b\gets\Call{BinarySearch}{\mathbf{J}_v[1:\ell_v],\,w}$
            \If{$b=\varnothing$} \textbf{continue} \Comment{$w\notin\mathcal{N}^{+}(v)$} \EndIf
            \State $s\gets s_{uv}+\mathbf{Q}_u[a]+\mathbf{Q}_v[b]$ \Comment{Eq.~\eqref{eq:clique}}
            \If{$s>\min_{\text{score}}R_t$} replace that minimum with $(s,w)$ \EndIf
        \EndFor
    \EndFor
    \State \textbf{block barrier}; write every $R_t$ into shared-memory buffer $\mathcal{M}$
    \State thread $0$: extract up to $T$ valid highest-scoring entries of $\mathcal{M}$ as the 3-cliques of $(u,v)$
\EndFor
\end{algorithmic}
\end{algorithm}

\subsubsection{Phase 1: Dataflow-Integrated Pivot Selection}
While constructing row $i$, the FFSOG block extracts its $m$ highest-scoring retained edges into $\mathcal{A}_i$ before the row leaves shared memory. Pivot selection therefore operates on at most $Nm$ scored edges instead of rescanning the complete SOG score matrix. Because each score is an integer common-neighbor count bounded by $L$, every block accumulates a local histogram of its candidate scores and merges it into a device-wide histogram $H$. A descending scan of $H$ finds the threshold score $\theta$; edges above $\theta$ are selected directly, and the remaining slots are filled from the $\theta$-bucket. If this boundary bucket contains more tied candidates than available slots, atomic arrival order determines which of them are retained; Sec.~\ref{sec:ablation} evaluates the resulting variation. Each selected pivot carries its score as $(u,v,s_{uv})$, so Phase~2 does not need to look up the pivot edge again. This compact selection avoids the dense score-matrix write, its subsequent read, and a host-side synchronization. 

The deployed path fuses SOG construction and histogram-based pivot selection into one cooperative kernel; an occupancy-bounded persistent grid enables grid-wide barriers between stages. On embedded devices without cooperative-launch support, the same stages execute as consecutive kernels on one stream, preserving the dataflow and output layout.

\subsubsection{Phase 2: Direct 3-Clique Enumeration}
For each pivot, Phase~2 enumerates valid third-vertex extensions and retains at most the $T$ highest-scoring 3-cliques. The ELL-SOG rows are written once to compact global memory because the two endpoints of a pivot may have been produced by different FFSOG blocks. One block is assigned to each pivot $(u,v)$, and its $B$ threads stride over the $\ell_u$ entries of $\mathbf{J}_u$, visiting every candidate $w$ exactly once. Each thread binary-searches its candidate in the sorted row $\mathbf{J}_v[1:\ell_v]$. Because row $v$ stores only columns greater than $v$, every candidate $w\le v$ misses, while every hit satisfies the canonical ordering $u<v<w$. A hit also returns the position $b$ and hence the edge score $\mathbf{Q}_v[b]$, allowing the thread to evaluate Eq.~\eqref{eq:clique} with three direct ELL-SOG reads. The total intersection work is $O(\ell_u\log\ell_v)$ per pivot rather than a scan over $N$ dense columns.

Each thread retains its $\kappa$ highest-scoring candidates in registers. Because $\kappa\ge T$, discarding a lower-ranked local candidate is safe: at least $T$ candidates processed by the same thread already have no lower scores, so the discarded candidate cannot enter the pivot's block-level top-$T$ set after merging. Local insertion requires neither atomics nor global scratch space. After enumeration, the thread-local lists are copied once to shared memory and merged, and up to $T$ highest-scoring triples are written to global memory for the pivot.

\subsubsection{Approximation and Bounded Cost}
FlashReg bounds two parts of the search. First, $L$ caps both the voter list used to compute a row of the SOG score matrix and the number of positive edges retained in that ELL-SOG row. Second, the row-local pool exposes only its $m$ highest-scoring retained edges to global pivot selection. These are independent restrictions: avoiding truncation by $L$ does not by itself guarantee that the top-$m$ pools contain every global top-$K$ pivot from the dense score matrix. FlashReg matches the dense search only if (i) $d_i\le L$ for every row, (ii) every edge in the dense global top-$K$ set appears in its row's top-$m$ pool, and (iii) pivot- and clique-boundary ties are resolved identically. Otherwise, it is a capacity-bounded approximation, whose empirical effect is reported in Sec.~\ref{sec:ablation}.

The capacities also bound the sparse search resources. The ELL-SOG and candidate pools use $O(NL+Nm)$ storage, and Phase~1 examines at most $Nm$ edges rather than all $N(N-1)/2$ cells of the dense SOG score matrix. For each of the $K$ pivots, Phase~2 searches one row of at most $L$ entries against another, giving $O(L\log L)$ work per pivot and $O(KL\log L)$ overall. At most $KT$ triples proceed to pose estimation. These bounds apply to the SOG score matrix and clique search; FlashReg still constructs the dense FOG adjacency matrix with $O(N^2)$ computation and storage.

% 基于数据流向优化的最大团搜索算法，目标：把 FOG→SOG→pivot 的多次 kernel 启动与中间结果的写回/重读全部消除，让数据"流过"而非"重建"，团搜索直接消费上游布局。具体加速步骤：
% 1) 单 cooperative kernel 融合三阶段：用 cg::grid_group + grid.sync() 在一个核内跑完
%    FOG→SOG→pivot top-K，取代原来"多次 launch + 每次把 SOG/候选写回全局显存再读回"的做法。
%    阶段划分：A(逐行算 SOG→padded 布局 + 候选池 + block 局部直方图) → B(局部直方图 flush 到全局)
%    → C(block 0 扫直方图定出 top-K 的阈值桶 tau) → D1/D2(按 >tau、==tau 收集 pivots)。
% 2) pivot top-K 在核内用"精确整数直方图桶选"完成，不回主机：因为 SOG 是小整数，
%    直方图桶选可给出与 torch::topk 完全一致的结果，省掉一次 GPU→host round-trip 和一次 O(N^2) 的 topk。
% 3) 候选池只有 [N, TOPM]，不是 N^2：pivot 选择只在这个小池上做，规模从 N^2 降到 N·TOPM。
% 4) 团搜索直接吃 FFSOG 的 padded 邻接，零重建：每个 block 处理一条 pivot 边 (u,v)，
%    线程跨 N(u) 并对每个 w 在 N(v) 上二分（交集 N(u)∩N(v)），打分 SOG(u,v)+SOG(u,w)+SOG(v,w)，
%    block 内归约出 per-pivot top-T。输入布局与 FFSOG 输出完全对齐，无需任何格式转换。
% 5) 可移植降级路径保持同一数据流：老 CUDA / Jetson（无 cooperativeLaunch）用 fused_sc2_pivot_compat，
%    拆成 4 次普通 launch，结果数值完全一致，仍不落地稠密矩阵——保证嵌入式平台上同样省内存。

\subsection{Pose Estimation, Verification, and Refinement}
\label{sec:pose}

The downstream stages follow TurboReg~\cite{turboreg} and are not a contribution of this work. Each retained 3-clique produces a pose hypothesis through a batched weighted-SVD (Kabsch) solver. The hypotheses are verified against the input correspondences, and the best pose is refined by iteratively reweighted SVD.

\section{Experiments}
\subsection{Experimental Setup}

We evaluate on the indoor 3DMatch~\cite{3dmatch} and 3DLoMatch~\cite{predator} benchmarks and the outdoor KITTI benchmark~\cite{kitti}, using both FPFH~\cite{fpfh} and FCGF~\cite{fcgf} correspondences. Let $N_{\max}$ denote the maximum number of input correspondences retained per pair, so the actual input size satisfies $N\le N_{\max}$. We use $N_{\max}{=}7000$ for indoor FPFH, $N_{\max}{=}6000$ for indoor FCGF, and $N_{\max}{=}5000$ for both descriptors on KITTI. A registration is successful when its rotation and translation errors are below $15^\circ$ and $30$\,cm indoors, or $5^\circ$ and $0.6$\,m on KITTI. We report registration recall (RR), mean rotation error (RE) and translation error (TE) over successful pairs, and per-pair runtime for benchmark evaluation.

We compare FlashReg with PointDSC~\cite{pointdsc}, VBReg~\cite{vbreg}, RANSAC~\cite{ransac}, FSAC-IA~\cite{fsacia}, TEASER++~\cite{teaser++}, SC2-PCR~\cite{sc2pcr}, MAC~\cite{mac}, FastMAC~\cite{fastmac}, and TurboReg~\cite{turboreg}. Unless otherwise stated, experiments run on the RTX~5090. Embedded-platform results use the Jetson AGX Xavier (Tab.~\ref{tab:experimental_platforms}).

For the indoor benchmarks, we set $\tau{=}0.012$\,m, $r{=}0.15$\,m, and $\tau_{\mathrm{in}}{=}0.1$\,m; the corresponding KITTI settings are $0.072$\,m, $1.0$\,m, and $0.6$\,m. Unless otherwise stated, we use $L{=}128$, $m{=}32$, $K{=}2000$, and $T{=}2$. The fused SOG/pivot kernel uses $256$ threads per block, while the clique kernel uses $B{=}64$ threads and a per-thread register capacity of $\kappa{=}4$.

Runtime covers the correspondence-to-pose pipeline from graph construction through pose refinement, after one warm-up and with device synchronization. Descriptor extraction, correspondence matching, and data transfers are excluded. All methods receive the same precomputed correspondences, which are initially resident in device memory for GPU methods and host memory for CPU methods. We test each baseline on the evaluation hardware using its public implementation; FlashReg and TurboReg additionally share the downstream stages described in Sec.~\ref{sec:pose}.
% TODO(yu): confirm baseline provenance for the record: TurboReg official code
% vs in-repo dense reference for Tables II-IV, repo commits for CPU baselines,
% and matching float precision; reviewers will probe the fairness setup.

\subsection{Benchmark Evaluation}

On the high-overlap 3DMatch benchmark (Tab.~\ref{tab:3dmatch}), FlashReg essentially matches TurboReg, reaching $84.01\%$ RR with FPFH and $93.41\%$ with FCGF. It also remains competitive with the strongest CPU clique method, trailing MAC by only $0.39$ percentage points under FPFH and $0.24$ points under FCGF. FlashReg reduces correspondence-to-pose latency from TurboReg's $10$\,ms to $4$--$5$\,ms and is over two orders of magnitude faster than the CPU clique solvers. Fig.~\ref{fig:speed_recall} highlights this accuracy--speed trade-off.

The same pattern holds on the more challenging 3DLoMatch benchmark (Tab.~\ref{tab:3dlomatch}). FlashReg obtains $39.25\%$ RR with FPFH and $59.12\%$ with FCGF, exceeding TurboReg by $0.06$ and $0.67$ percentage points, respectively, while remaining within $1.74$ and $0.90$ points of MAC. Its latency remains $4$--$5$\,ms per pair, more than $2\times$ faster than TurboReg and about $200\times$ faster than MAC. Among successful registrations, FlashReg also reduces translation error relative to TurboReg for both descriptors.

On the outdoor KITTI benchmark (Tab.~\ref{tab:kitti}), FlashReg matches TurboReg's recall exactly, with $97.83\%$ under FPFH and $97.66\%$ under FCGF, while outperforming all evaluated CPU methods. It reduces TurboReg's $11$--$12$\,ms latency to $4$--$5$\,ms. Across all three benchmarks, FlashReg therefore preserves the accuracy of GPU clique-based registration while consistently delivering a $2$--$3\times$ speedup.

\subsection{Embedded Platform Evaluation}

We evaluate embedded deployment by running FlashReg and TurboReg on the Jetson AGX Xavier. On 3DMatch, we sweep $N_{\max}$ for pivot budgets $K\in\{1000,1500,2000\}$. The Jetson operates in its maximum-power mode with clocks locked by \texttt{nvpmodel -m 0} and \texttt{jetson\_clocks}. For every $(N_{\max},K)$ setting, both methods process the same complete set of 1,623 pairs. The reported energy is the total GPU-rail energy for that run, obtained by sampling \texttt{tegrastats} every 20\,ms and integrating the sum of rails whose reported names contain ``GPU''; idle power is not subtracted. Peak allocated tensor memory is the maximum warm-up-excluded live-tensor allocation reported by the PyTorch CUDA allocator and includes both input correspondence tensors and algorithm buffers; it excludes the CUDA context, driver memory, and allocations outside the PyTorch allocator.

\begin{figure*}[!t]
    \centering
    \includegraphics[width=\linewidth]{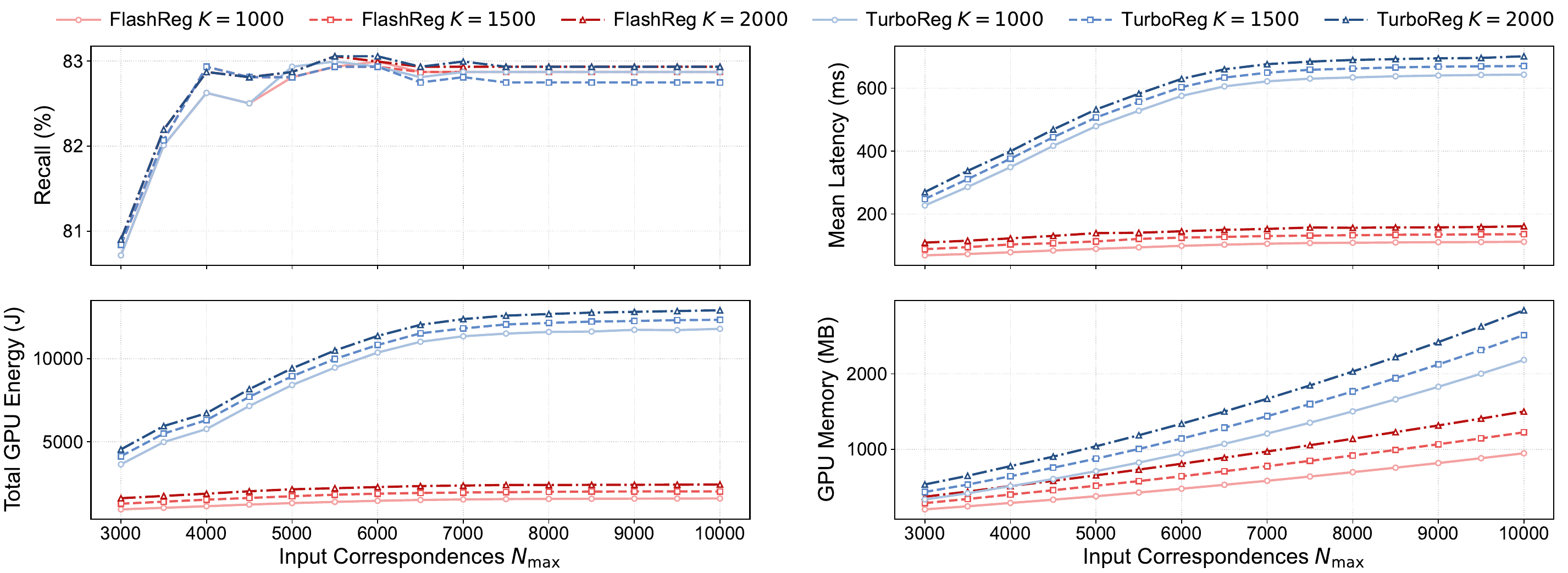}
    \caption{Recall, latency, total GPU-rail energy, and peak PyTorch-allocated tensor memory on the Jetson AGX Xavier (3DMatch) as $N_{\max}$ and $K$ vary. Energy is integrated over the complete 1,623-pair run. The ``GPU Memory'' panel denotes this allocator metric, not total device-memory use; the recall axis is truncated to expose small differences.}
    \label{fig:jetson_performance}
\end{figure*}

Fig.~\ref{fig:jetson_performance} reports recall, latency, total GPU-rail energy, and peak allocated tensor memory across these settings. FlashReg tracks TurboReg's recall while using about half as much peak allocated tensor memory and consistently less measured energy. Eliminating the dense SOG score matrix substantially reduces tensor-memory growth over the evaluated $N_{\max}$ range, although both methods still retain the dense FOG adjacency matrix. This reduced growth is particularly valuable on the Jetson's unified-memory architecture.

\subsection{Ablation Study}
\label{sec:ablation}

Tab.~\ref{tab:ablation} isolates the contribution of each FlashReg component on 3DMatch with FPFH. FastFOG and FastSOG accelerate their respective stages from $1.64$ to $0.19$,ms ($8.6\times$) and from $3.56$ to $0.84$,ms ($4.2\times$). Combined, they reduce the total latency from $11.40$ to $7.22$,ms, with only a $0.13$ percentage-point change in RR. The dataflow design further reduces clique enumeration from $0.54$ to $0.04$,ms and reorganizes SOG construction and pivot selection; the fused implementation completes these operations in $0.42$,ms. With all components enabled, the split and fused pipelines achieve total latencies of $5.16$ and $5.11$,ms, respectively. Overall, the deployed fused pipeline delivers a $2.23\times$ end-to-end speedup ($55.2\%$ latency reduction) over the dense reference while maintaining $84.01\%$ RR, only $0.16$ percentage points lower than the reference.

% \usepackage{graphicx}
\begin{table}[t]
    \centering
    \caption{Hardware and software configurations used in the experiments.}
    \label{tab:experimental_platforms}
    \scriptsize
    \begin{tabular}{@{}lll@{}}
        \toprule
        \textbf{Device} & \textbf{Desktop PC}     & \textbf{Jetson AGX Xavier} \\
        \midrule
        CPU          & AMD Ryzen 7 9850X3D     & NVIDIA Carmel           \\
        GPU          & NVIDIA GeForce RTX 5090 & 512-core NVIDIA Volta    \\
        GPU memory & 32\,GB (VRAM)   & 16\,GB (unified)          \\
        CUDA version & 13.0                    & 10.2                     \\
        % TODO(yu): add GPU driver and JetPack/L4T versions before submission.
        \bottomrule
    \end{tabular}
\end{table}

% \usepackage{multirow}
\begin{table}[t]
\caption{Performance comparison of different registration methods on the 3DMatch dataset.}
\label{tab:3dmatch}
\setlength{\tabcolsep}{5pt}
\resizebox{\columnwidth}{!}{%
\begin{tabular}{@{}lcccccccc@{}}
        \toprule
        \multirow{2}{*}{Method} & \multicolumn{4}{c}{FPFH} & \multicolumn{4}{c}{FCGF} \\
        \cmidrule(lr){2-5} \cmidrule(lr){6-9}
         & RR (\%) & RE (deg) & TE (cm) & Time (s) & RR (\%) & RE (deg) & TE (cm) & Time (s) \\
        \midrule
        \multicolumn{9}{@{}l}{\textit{Learning-based}} \\
        \textbf{PointDSC*} & 77.51 & 2.07 & 6.45 & 0.035 & 92.91 & 2.04 & 6.52 & 0.034 \\
        \textbf{VBReg*} & 79.54 & 2.27 & 7.01 & 0.072 & 93.04 & 2.07 & 6.57 & 0.079 \\
        \midrule
        \multicolumn{9}{@{}l}{\textit{Learning-free}} \\
        \textbf{RANSAC-10k} & 70.10 & 4.13 & 11.63 & 2.533 & 81.80 & 3.71 & 11.12 & 1.760 \\
        \textbf{FSAC-IA} & 69.58 & 4.13 & 12.05 & 0.247 & 82.62 & 3.79 & 11.34 & 0.215 \\
        \textbf{TEASER++} & 80.90 & 2.30 & 7.13 & 0.361 & 88.54 & 2.23 & 7.46 & 1.039 \\
        \textbf{SC2-PCR} & 83.80 & 2.12 & 10.66 & 0.776 & 93.16 & 2.06 & 6.53 & 0.783 \\
        \textbf{MAC} & 84.40 & 2.09 & 6.73 & 1.015 & 93.65 & 2.03 & 6.52 & 0.911 \\
        \textbf{FastMAC} & 82.50 & 2.13 & 6.75 & 0.545 & 92.98 & 2.01 & 6.49 & 0.521 \\
        \textbf{TurboReg*} & 84.17 & 2.04 & 6.57 & 0.010 & 93.59 & 2.03 & 6.46 & 0.010 \\
        \textbf{FlashReg*} & 84.01 & 2.04 & 6.56 & \textbf{0.004} & 93.41 & 2.03 & 6.48 & 0.005 \\
        \bottomrule
    \end{tabular}%
}

{\footnotesize $^{*}$\,GPU-based method; the others run on CPU. All experiments were conducted on the desktop PC (RTX 5090).}
\end{table}

% \usepackage{multirow}
\begin{table}[h]
\caption{Performance comparison of different registration methods on the 3DLoMatch dataset.}
\label{tab:3dlomatch}
\setlength{\tabcolsep}{5pt}
\resizebox{\columnwidth}{!}{%
\begin{tabular}{@{}lcccccccc@{}}
        \toprule
        \multirow{2}{*}{Method} & \multicolumn{4}{c}{FPFH} & \multicolumn{4}{c}{FCGF} \\
        \cmidrule(lr){2-5} \cmidrule(lr){6-9}
         & RR (\%) & RE (deg) & TE (cm) & Time (s) & RR (\%) & RE (deg) & TE (cm) & Time (s) \\
        \midrule
        \multicolumn{9}{@{}l}{\textit{Learning-based}} \\
        \textbf{PointDSC*} & 27.51 & 4.24 & 10.42 & 0.032 & 55.14 & 3.85 & 10.45 & 0.033 \\
        \textbf{VBReg*} & 33.58 & 4.38 & 10.77 & 0.074 & 57.94 & 3.80 & 10.75 & 0.074 \\
        \midrule
        \multicolumn{9}{@{}l}{\textit{Learning-free}} \\
        \textbf{RANSAC-10k} & 5.86 & 6.21 & 14.13 & 2.257 & 27.24 & 5.08 & 13.62 & 1.770 \\
        \textbf{FSAC-IA} & 6.26 & 5.91 & 14.62 & 0.235 & 27.27 & 5.03 & 14.01 & 0.201 \\
        \textbf{TEASER++} & 34.03 & 4.30 & 10.68 & 0.476 & 48.85 & 4.29 & 11.92 & 0.516 \\
        \textbf{SC2-PCR} & 36.16 & 4.23 & 10.66 & 0.598 & 57.89 & 3.78 & 10.58 & 0.539 \\
        \textbf{MAC} & 40.99 & 4.03 & 10.75 & 0.785 & 60.02 & 3.73 & 10.55 & 1.091 \\
        \textbf{FastMAC} & 38.24 & 4.13 & 10.82 & 0.461 & 57.61 & 3.71 & 10.50 & 0.603 \\
        \textbf{TurboReg*} & 39.19 & 4.49 & 15.34 & 0.011 & 58.45 & 2.49 & 9.83 & 0.011 \\
        \textbf{FlashReg*} & 39.25 & 4.04 & 10.34 & \textbf{0.004} & 59.12 & 2.49 & 7.96 & 0.005 \\
        \bottomrule
    \end{tabular}%
}

{\footnotesize $^{*}$\,GPU-based method; the others run on CPU. All experiments were conducted on the desktop PC (RTX 5090).}
\end{table}

% \usepackage{multirow}
\begin{table}[h]
% TODO(yu): the PointDSC row below is almost certainly a copy-paste of the
% 3DMatch row (RR 77.51 identical; RE ~2 deg is implausible on KITTI, and the
% MAC paper reports PointDSC KITTI-FPFH RR 98.92 / RE 0.38 / TE 8.35). Re-run
% or re-source these numbers before submission.
\caption{Performance comparison of different registration methods on the KITTI dataset.}
\label{tab:kitti}
\setlength{\tabcolsep}{5pt}
\resizebox{\columnwidth}{!}{%
\begin{tabular}{@{}lcccccccc@{}}
        \toprule
        \multirow{2}{*}{Method} & \multicolumn{4}{c}{FPFH} & \multicolumn{4}{c}{FCGF} \\
        \cmidrule(lr){2-5} \cmidrule(lr){6-9}
         & RR (\%) & RE (deg) & TE (cm) & Time (s) & RR (\%) & RE (deg) & TE (cm) & Time (s) \\
        \midrule
        \multicolumn{9}{@{}l}{\textit{Learning-based}} \\
        \textbf{PointDSC*} & 97.29 & 0.44 & 9.56 & 0.047 & 98.74 & 0.21 & 11.01 & 0.046 \\
        \textbf{VBReg*} & 97.83 & 0.43 & 9.48 & 0.054 & 98.92 & 0.28 & 10.92 & 0.055 \\
        \midrule
        \multicolumn{9}{@{}l}{\textit{Learning-free}} \\
        \textbf{RANSAC-10k} & 47.21 & 1.57 & 28.69 & 1.985 & 80.36 & 0.60 & 20.42 & 0.433 \\
        \textbf{FSAC-IA} & 47.21 & 1.60 & 30.01 & 0.254 & 81.02 & 0.60 & 19.94 & 0.068 \\
        \textbf{TEASER++} & 93.17 & 0.50 & 17.98 & 0.448 & 94.64 & 0.28 & 13.69 & 2.215 \\
        \textbf{SC2-PCR} & 94.56 & 0.28 & 20.08 & 1.715 & 96.56 & 0.36 & 8.18 & 1.509 \\
        \textbf{MAC} & 97.11 & 0.39 & 8.49 & 0.648 & 97.48 & 0.36 & 8.14 & 0.201 \\
        \textbf{FastMAC} & 97.12 & 0.43 & 9.06 & 0.348 & 97.30 & 0.43 & 9.06 & 0.051 \\
        \textbf{TurboReg*} & 97.83 & 0.41 & 8.13 & 0.011 & 97.66 & 0.36 & 7.72 & 0.012 \\
        \textbf{FlashReg*} & 97.83 & 0.41 & 8.13 & \textbf{0.004} & 97.66 & 0.35 & 7.67 & 0.005 \\
        \bottomrule
    \end{tabular}%
}

{\footnotesize $^{*}$\,GPU-based method; the others run on CPU. All experiments were conducted on the desktop PC (RTX 5090).}
\end{table}

% Ablation of the on-chip modules (per-stage GPU time). Requires \usepackage{booktabs} and \usepackage{amssymb} (\checkmark).
\begin{table*}[!t]
\centering
\fontsize{7.5}{9}\selectfont
\setlength{\tabcolsep}{5pt}
\caption{Component ablation on 3DMatch with FPFH. Total includes verification, refinement, and pose solving. Full (fused) is the deployed pipeline. All ablation experiments were conducted on the desktop PC (RTX 5090). }
\label{tab:ablation}
\begin{tabular}{@{}l cccc c ccccc@{}}
\toprule
& \multicolumn{4}{c}{Modules enabled} & & \multicolumn{5}{c}{Per-stage GPU time (ms)} \\
\cmidrule(lr){2-5} \cmidrule(lr){7-11}
Configuration & FastFOG & FastSOG & Dataflow & Pivot & RR (\%) & FOG & SOG & Pivot & Clique & Total \\
\midrule
Dense reference                &            &            &            &        & 84.17 & 1.64 & 3.56 & 1.21 & 0.55 & 11.40 \\
FastFOG                        & \checkmark &            &            &        & 84.17 & 0.19 & 3.56 & 1.21 & 0.55 &  9.95 \\
FastSOG                        &            & \checkmark &            &        & 84.04 & 1.64 & 0.84 & 1.19 & 0.54 &  8.69 \\
FastFOG\,$+$\,FastSOG          & \checkmark & \checkmark &            &        & 84.04 & 0.19 & 0.84 & 1.19 & 0.54 &  7.22 \\
FastSOG\,$+$\,Dataflow (fused) &            & \checkmark & \checkmark & fused  & 84.01 & 1.64 & 0.42 &   --   & 0.04 &  6.55 \\
FastSOG\,$+$\,Dataflow (split) &            & \checkmark & \checkmark & split  & 84.04 & 1.64 & 0.39 & 0.09 & 0.04 &  6.60 \\
Full (split)                   & \checkmark & \checkmark & \checkmark & split  & 84.04 & 0.19 & 0.39 & 0.09 & 0.04 & 5.16 \\
Full (fused)                   & \checkmark & \checkmark & \checkmark & fused  & 84.01 & 0.19 & 0.41 &   --   & 0.04 & \textbf{5.11} \\
\bottomrule
\end{tabular}
\end{table*}

\section{Conclusion}
This work presented FlashReg, a GPU-native 3-clique correspondence-to-pose estimator that avoids materializing the dense SOG score matrix and eliminates redundant data movement between graph construction and search. Its consumer-oriented ELL-SOG and candidate pools preserve recall comparable to TurboReg while reducing estimator latency by $2$--$3\times$. On the Jetson AGX Xavier, the same design also lowers peak allocated tensor memory and measured GPU-rail energy, making FlashReg a practical backend for onboard registration pipelines. Future work will extend the dataflow design to higher-order geometric reasoning and integrate it with complete perception systems.

\bibliographystyle{unsrt}
\bibliography{reference}

\end{document}